\documentclass[11pt,a4paper]{article}
\usepackage[hyperref]{emnlp2020}
\usepackage{times}
\usepackage{latexsym}

\usepackage{microtype}

\usepackage{xcolor,colortbl}
\usepackage{url}
\usepackage{graphicx}
\usepackage{multirow,array,booktabs}
\usepackage{etoolbox}
\usepackage{ragged2e}
\usepackage{amsmath} 
\usepackage{amssymb}
\usepackage{enumitem}
\usepackage{epigraph} 
\usepackage{enumitem}
\usepackage{hang}
\usepackage{tablefootnote}

\aclfinalcopy 

\title{Contextual Modulation for Relation-Level Metaphor Identification}

\author{Omnia Zayed, John P. McCrae, Paul Buitelaar \\
  Insight SFI Research Centre for Data Analytics\\
  Data Science Institute \\
  National University of Ireland Galway \\
  IDA Business Park, Lower Dangan, Galway, Ireland\\
  \texttt{\{omnia.zayed, john.mccrae, paul.buitelaar\}@insight-centre.org} \\}

\date{}

\begin{document}
\maketitle
\begin{abstract}
Identifying metaphors in text is very challenging and requires comprehending the underlying comparison. The automation of this cognitive process has gained wide attention lately. However, the majority of existing approaches concentrate on word-level identification by treating the task as either single-word classification or sequential labelling without explicitly modelling the interaction between the metaphor components. On the other hand, while existing relation-level approaches implicitly model this interaction, they ignore the context where the metaphor occurs. In this work, we address these limitations by introducing a novel architecture for identifying relation-level metaphoric expressions of certain grammatical relations based on contextual modulation. In a methodology inspired by works in visual reasoning, our approach is based on conditioning the neural network computation on the deep contextualised features of the candidate expressions using feature-wise linear modulation. We demonstrate that the proposed architecture achieves state-of-the-art results on benchmark datasets. The proposed methodology is generic and could be applied to other textual classification problems that benefit from contextual interaction.
\end{abstract}

\section{Introduction}
\label{sec:intro}
Despite its fuzziness, metaphor is a fundamental feature of language that defines the relation between how we understand things and how we express them~\cite{cameron_low:1999}. A metaphor is a figurative device containing an implied mapping between two conceptual domains. These domains are represented by its two main components, namely the tenor (target domain) and the vehicle (source domain)~\cite{end:1986:grounds}. 
According to the conceptual metaphor theory (CMT) of~\citet{LakoffJohnson80}, which we adopt in this work, a concept such as \textit{``liquids''} (source domain/vehicle) can be borrowed to express another such as \textit{``emotions''} (target domain/tenor) by exploiting single or common properties. Therefore, the conceptual metaphor \textit{``Emotions are Liquids''} can be manifested through the use of linguistic metaphors such as \textit{``pure love''}, \textit{``stir excitement''} and \textit{``contain your anger''}. The interaction between the target and the source concepts of the expression is important to fully comprehend its metaphoricity.  

Over the last couple of years, there has been an increasing interest towards metaphor processing and its applications, either as part of natural language processing (NLP) tasks such as machine translation~\cite{koglin:2019:metaphor_translation}, text simplification~\cite{wolska_clausen:2017:metaphor_text_simplification,clausen_nastase:2019:metaphor_text_simplification} and sentiment analysis~\cite{Rentoumi:2012:metaphor_sentiment_analysis} or in more general discourse analysis use cases such as in analysing political discourse~\cite{Charteris-Black:2011}, financial reporting~\cite{Ho_Cheng:2016} and health communication~\cite{Semino:2018}. 

Metaphor processing comprises several tasks including identification, interpretation and cross-domain mappings. Metaphor identification is the most studied among these tasks. It is concerned with detecting the metaphoric words or expressions in the input text and could be done either on the sentence, relation or word levels. The difference between these levels of processing is extensively studied in~\cite{Zayed:2020}. Identifying metaphors on the word-level could be treated as either \textit{sequence labelling} by deciding the metaphoricity of each word in a sentence given the context or \textit{single-word classification} by deciding the metaphoricity of a targeted word. On the other hand, relation-level identification looks at specific grammatical relations such as the \textit{dobj} or \textit{amod} dependencies and checks the metaphoricity of the verb or the adjective given its association with the noun. In relation-level identification, both the source and target domain words (the tenor and vehicle) are classified either as a metaphoric or literal expression, whereas in word-level identification only the source domain words (vehicle) are labelled. These levels of analysis (paradigms) are already established in literature and adopted by previous research in this area as will be explained in Section~\ref{sec:related_work}. The majority of existing approaches, as well as the available datasets, pertaining to metaphor processing focus on the metaphorical usage of verbs and adjectives either on the word or relation levels. This is because these syntactic types exhibit metaphoricity more frequently than others according to corpus-based analysis~\cite{cameron:2003,ShutovaTeufel:2010}.

Although the main focus of both the relation-level and word-level metaphor identification is discerning the metaphoricity of the vehicle (source domain words), the interaction between the metaphor components is less explicit in word-level analysis either when treating the task as sequence labelling or single-word classification. Relation-level analysis could be viewed as a deeper level analysis that captures information that is not captured on the word-level through modelling the influence  of  the  tenor  (e.g.noun) on the vehicle (e.g. verb/adjective). There will be reasons that some downstream tasks would prefer to have such information (i.e. explicitly marked relations), among these tasks are metaphor interpretation and cross-domain mappings. Moreover, employing the wider context around the expression is essential to improve the identification process.

This work focuses on relation-level metaphor identification represented by verb-noun and adjective-noun grammar relations. We propose a novel approach for context-based textual classification that utilises affine transformations. In order to integrate the interaction of the metaphor components in the identification process, we utilise affine transformation in a novel way to condition the neural network computation on the contextualised features of the given expression. The idea of affine transformations has been used in NLP-related tasks such as visual question-answering~\cite{deVries:2017_visualQA}, dependency parsing~\cite{Dozat-Manning:2017:biaffine-parsing}, semantic role labelling~\cite{cai:2018:biaffine-srl}, coreference resolution~\cite{zhang:2018:biaffine-coreference}, visual reasoning~\cite{perez:2018_film} and lexicon features integration~\cite{margatina:2019-attention}.  

Inspired by the works on visual reasoning, we use the candidate expression of certain grammatical relations, represented by deep contextualised features, as an auxiliary input to modulate our computational model. Affine transformations can be utilised to process one source of information in the context of another. In our case, we want to integrate: 1) the deep contextualised-features of the candidate expression (represented by ELMo sentence embeddings) with 2) the syntactic/semantic features of a given sentence. Based on this task, affine transformations have a similar role to attention but with more parameters, which allows the model to better exploit context. Therefore, it could be regarded as a form of a more sophisticated attention. Whereas the current ``straightforward'' attention models are overly simplistic, our model prioritises the contextual information of the candidate to discern its metaphoricity in a given sentence.

Our proposed model consists of an affine transform coefficients generator that captures the meaning of the candidate to be classified, and a neural network that encodes the full text in which the candidate needs to be classified. We demonstrate that our model significantly outperforms the state-of-the-art approaches on existing relation-level benchmark datasets. The unique characteristics of tweets and the availability of Twitter data motivated us to identify metaphors in such content. Therefore, we evaluate our proposed model on a newly introduced dataset of tweets~\cite{Zayed:2019} annotated for relation-level metaphors. 

\section{Related Work}
\label{sec:related_work}
Over the last decades, the focus of computational metaphor identification has shifted from rule-based~\cite{fass:91} and knowledge-based approaches~\cite{krishnakumaran_zhu:2007,Wilks:2013} to statistical and machine learning approaches including supervised~\cite{Gedigian:2006,Turney:2011,dunn:2013:cicling, dunn:2013:META4NLP,Tsvetkov:2013:Meta4NLP,Hovy:2013,Mohler:2013:Meta4NLP,klebanov:2014:different,Bracewell:2014,jang:2015,gargett-barnden:2015:META4NLP,rai:2016:Meta4NLP,bulat:2017,Koeper:2017}, semi-supervised~\cite{Birke:2006,Shutova:2010Ident,Zayed:2018} and unsupervised methods~\cite{ShutovaAndSun:2013:graphcluster,Heintz:2013,Strzalkowski:2013:Meta4NLP}. These approaches employed a variety of features to design their models. With the advances in neural networks, the focus started to shift towards employing more sophisticated models to identify metaphors. This section focuses on current research that employs neural models for metaphor identification on both word and relation levels. 

\begin{description}[topsep=1pt,itemsep=0pt,leftmargin=0pt]
\item[\textbf{Word-Level Processing:}] \citet{Do-Dinh_Gurevych:2016} were the first to utilise a neural architecture to identify metaphors. They approached the problem as sequence labelling where a traditional fully-connected feed-forward neural network is trained using pre-trained word embeddings. The authors highlighted the limitation of this approach when dealing with short and noisy conversational texts.
As part of the NAACL 2018 Metaphor Shared Task~\cite{leong:2018:report}, many researchers proposed neural models that mainly employ LSTMs~\cite{HochSchm:97:LSTM} with pre-trained word embeddings to identify metaphors on the word-level. The best performing systems are: THU NGN \cite{wu:2018}, OCOTA \cite{bizzoni:2018} and bot.zen \cite{stemle:2018}.
\citet{gao:2018} were the first to employ the deep contextualised word representation ELMo~\cite{Peters:2018}, combined with pre-trained GloVe~\cite{pennington:2014glove} embeddings to train bidirectional LSTM-based models. The authors introduced a sequence labelling model and a single-word classification model for verbs. They showed that incorporating the context-dependent representation of ELMo with context-independent word embeddings improved metaphor identification.
\citet{mu:2019} proposed a system that utilises a gradient boosting decision tree classifier. Document embeddings were employed in an attempt to exploit wider context to improve metaphor detection in addition to other word representations including GLoVe, ELMo and skip-thought~\cite{kiros:2015:skip_thought}. 
\citet{mao:2018-word,mao:2019_end} explored the idea of selectional preferences violation~\cite{wilks:78} in a neural architecture to identify metaphoric words. Mao's proposed approaches emphasised the importance of the context to identify metaphoricity by employing context-dependent and context-independent word embeddings. \citet{mao:2019_end} also proposed employing multi-head attention to compare the targeted word representation with its context. 
An interesting approach was introduced by~\citet{dankers:2019:interplay-emotions} to model the interplay between metaphor identification and emotion regression. The authors introduced multiple multi-task learning techniques that employ hard and soft parameter sharing methods to optimise LSTM-based and BERT-based models.

\item[\textbf{Relation-Level Processing:}] \citet{Shutova:2016} focused on identifying the metaphoricity of adjective/verb-noun pairs. This work employed multimodal embeddings of visual and linguistic features. Their model employs the cosine similarity of the candidate expression components based on word embeddings to classify metaphors using an optimised similarity threshold.
\citet{rei:2017} introduced a supervised similarity network to detect adjective/verb-noun metaphoric expressions. Their system utilises word gating, vector representation mapping and a weighted similarity function. Pre-trained word embeddings and attribute-based embeddings~\cite{bulat:2017} were employed as features. This work explicitly models the interaction between the metaphor components. Gating is used to modify the vector of the verb/adjective based on the noun, however the surrounding context is ignored by feeding only the candidates as input to the neural model which might lead to loosing important contextual information.

\item[\textbf{Limitations:}] As discussed, the majority of previous works adopted the word-level paradigm to identify metaphors in text. 
The main distinction between the relation-level and the word-level paradigms is that the former makes the context more explicit than the latter through providing information about not only where the metaphor is in the sentence but also how its components come together through hinting at the relation between the tenor and the vehicle. 
\citet{stowe:2018:figlang} showed that the type of syntactic construction a verb occurs in influences its metaphoricity. On the other hand, existing relation-level approaches~\cite{Tsvetkov:2014,Shutova:2016,bulat:2017,rei:2017} ignore the context where the expression appears and only classify a given syntactic construction as metaphorical or literal. Studies showed that the context surrounding a targeted expression is important to discern its metaphoricity and fully grasp its meaning~\cite{mao:2018-word,mu:2019}. Therefore, current relation-level approaches will only be able to capture commonly used conventionalised metaphors. In this work, we address these limitations by introducing a novel approach to textual classification which employs contextual information from both the targeted expression under study and the wider context surrounding it. 

\end{description}

\section{Proposed Approach}
\label{sec:proposed_approach}

Feature-wise transformation techniques such as feature-wise linear modulation (FiLM) have been recently employed in many applications showing improved performance. They became popular in image processing applications such as image style transfer~\cite{Dumoulin:2017_style}; then they found their way into multi-modal tasks, specifically visual question-answering \cite{deVries:2017_visualQA,perez:2018_film}. They also have been shown to be effective approaches for relational problems as mentioned in Section~\ref{sec:intro}. The idea behind FiLM is to condition the computation carried out by a neural model on the information extracted from an auxiliary input in order to capture the relationship between multiple sources of information \cite{dumoulin:2018_feature-wise}. 

Our approach adopts Perez's \citeyearpar{perez:2018_film} formulation of FiLM on visual reasoning for metaphor identification. In visual reasoning, image-related questions are answered by conditioning the image-based neural network (visual pipeline) on the question context via a linguistic pipeline. In metaphor identification, we can consider that the image in our case is the sentence that has a metaphoric candidate and the auxiliary input is the linguistic interaction between the components of the candidate itself. This will allow us to condition the computation of a sequential neural model on the contextual information of the candidate and leverage the feature-wise interactions between the conditioning representation and the conditioned network. To the best of our knowledge, we are the first to propose such contextual modulation for textual classification in general and for metaphor identification specifically.

Our proposed architecture consists of \textit{a contextual modulation} pipeline and \textit{a metaphor identification linguistic} pipeline as shown in Figure~\ref{fig:model}. The input to the contextual modulator is the deep contextualised representation of the candidate expression under study (which we will refer to as targeted expression\footnote{Targeted expressions are already annotated in the dataset and initially obtained either manually or automatically using a dependency parser as will be described in Section~\ref{sec:datasets}.}) to capture the interaction between its components. The linguistic pipeline employs an LSTM encoder which produces a contextual representation of the provided sentence where the targeted expression appeared. The model is trained end-to-end to identify relation-level metaphoric expressions focusing on verb-noun and adjective-noun grammatical relations. Our model takes as input a sentence (or a tweet) and a targeted expression of a certain syntactic construction and identifies whether the candidate in question is used metaphorically or literally by going through the following steps:

\begin{figure*}[!htb]
	\begin{center}
		\includegraphics[width=\textwidth]{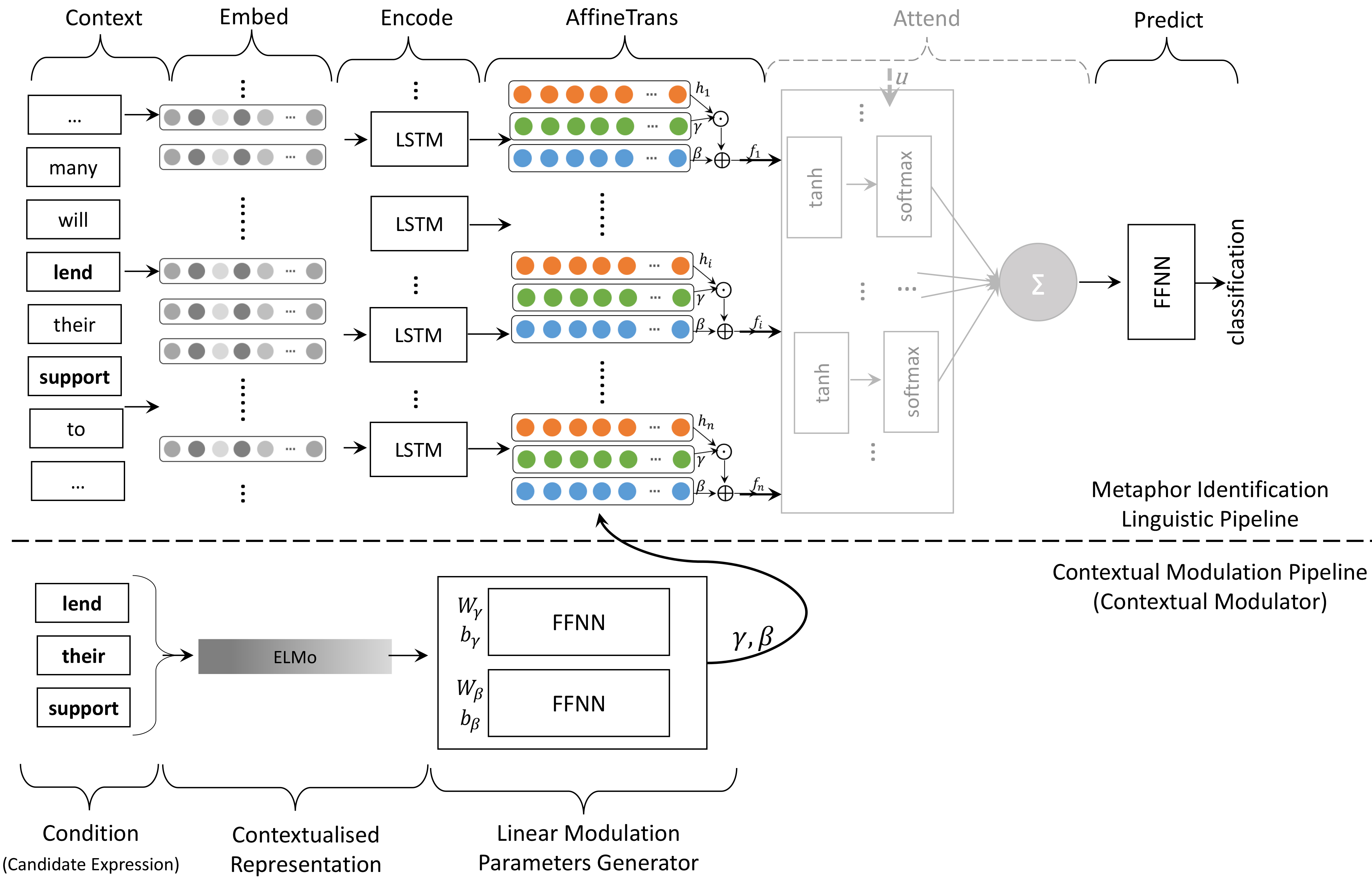} 
		\caption{\label{fig:model} The proposed framework for relation-level metaphor identification showing the contextual modulation in detail. The attention process is greyed out as we experimented with and without it.}
	\end{center}
\end{figure*}

\begin{description}[topsep=1pt,itemsep=0pt,leftmargin=0pt]
\item[\textbf{Condition:}] In this step the targeted expression is used as the auxiliary input to produce a conditioning representation. We first embed each candidate of verb-direct object pairs\footnote{We do the same for subject-verb and adjective-noun pairs but, for simplicity, we demonstrate the process with verb-direct object pairs.} $(v,n)$ using ELMo sentence embeddings to learn context-dependent aspects of word meanings $c_{vn}$. We used the 1,024-dimensional ELMo embeddings pre-trained on the One Billion Word benchmark corpus~\cite{chelba:2014}. The sentence embeddings of the targeted expression are then prepared by implementing an embeddings layer that loads these pre-trained ELMo embeddings from the TensorFlow Hub\footnote{\url{https://www.tensorflow.org/hub}}. The layer takes in the raw text of the targeted expression and outputs a fixed mean-pooled vector representation of the input as the contextualised representation. This representation is then used as an input to the main component of this step, namely a contextual modulator. The contextual modulator consists of a fully-connected feed-forward neural network (FFNN) that produces the conditioning parameters (i.e. the shifting and scaling coefficients) that will later modulate the linguistic pipeline computations. Given that $c_{vn}$ is the conditioning input then the contextual modulator outputs $\gamma$ and $\beta$, the context-dependent scaling and shifting vectors, as follows:
\begin{equation}
\begin{aligned}
    \gamma(c_{vn}) = W_{\gamma}c_{vn} + b_{\gamma},
   \\ 
    \beta(c_{vn}) = W_{\beta}c_{vn} + b_{\beta}
\end{aligned}
\end{equation}
where \textit{$W_\gamma$, $W_\beta$,  $b_\gamma$, $b_\beta$} are learnable parameters. 

\item[\textbf{Embed:}] Given a labelled dataset of sentences, the model begins by embedding the tokenised sentence $S$ of words $w_1, w_2,...,w_n$, where $n$ is the number of words in $S$, into vector representations using GloVe embeddings. We used the uncased 200-dimensional GloVe embeddings pre-trained on $\sim$2 billion tweets and contains 1.2 million words.

\item[\textbf{Encode:}] The next step is to train a neural network with the obtained embeddings. Since context is important for identifying metaphoricity, sentence encoder is a sensible choice. We use an LSTM sequence model to obtain a contextual representation which summarises the syntactic and semantic features of the whole sentence. The output of the LSTM is a sequence of hidden states $h_1, h_2, ..., h_n$, where $h_i$ is the hidden state at the $i^{th}$ time-step.

\item[\textbf{Feature-wise Transformation:}] In this step, an affine transformation layer, hereafter \textit{AffineTrans} layer, applies a feature-wise linear modulation to its inputs, which are: 1) the hidden states from the encoding step; 2) the scaling and shifting parameters from the conditioning step. By feature-wise, we mean that scaling and shifting are applied to each encoded vector for each word in the sentence. 
\begin{align}
f(h_i,c_{vn}) = \gamma(c_{vn}) \odot h_i + \beta(c_{vn})
\end{align}

\item[\textbf{Attend:}] Recently, attention mechanisms have become useful to select the most important elements in a given representation while minimising information loss. In this work, we employ an attention layer based on the mechanism presented in~\cite{lin:2017:attent}. It takes the output from the \textit{AffineTrans} layer as an input in addition to a randomly initialised weight matrix $W$, a bias vector $b$ and a learnable context vector $u$ to produce the attended output as follows:
\begin{align}
    e_{i} = tanh(W f_i + b)
    \label{eq4}\\
   \alpha_{i} = softmax(u e_{i})
   \label{eq5}\\
    r = \sum^{n}_{i=1} \alpha_{i} f_i 
    \label{eq6}
\end{align}
Our model is trained and evaluated with and without the attention mechanism in order to differentiate between the effect of the feature modulation and the attention on the model performance.

\item[\textbf{Predict:}] The last step is to make the final prediction using the output from the previous step (attended output in case of using attention or the \textit{AffineTrans} layer output in case of skipping it). We use a fully-connected feed-forward layer with a sigmoid activation that returns a single (binary) class label to identify whether the targeted expression is metaphoric or not.
\end{description}

\section{Datasets}
\label{sec:datasets}
The choice of annotated dataset for training the model and evaluating its performance is determined by the level of metaphor identification. 
Given the distinction between the levels of analysis, approaches addressing the task on the word-level are not fairly comparable to relation-level approaches since each task addresses metaphor identification differently. Therefore, the tradition of previous work in this area is to compare approaches addressing the task on the same level against each other on level-specific annotated benchmark datasets~\cite{Zayed:2020}.

Following prior work in this area and in order to compare the performance of our proposed approach with other relation-level metaphor identification approaches, we utilise available annotated datasets that support this level of processing. The existing datasets are either originally prepared to directly support relation-level processing such as the TSV~\cite{Tsvetkov:2014} dataset and the Tweets dataset by~\citet{Zayed:2019} or adapted from other word-level benchmark datasets to suit relation-level processing such as the adaptation of the benchmark datasets TroFi~\cite{Birke:2006} and VU Amsterdam metaphor corpus (VUAMC)~\cite{Steen:2010} by~\citet{Zayed:2020} and the adaptation of the MOH~\cite{Mohammad:2016} dataset by~\citet{Shutova:2016}. Due to space limitation, we include in Appendix~\ref{sec:appendix_a}: 1) examples of annotated instances from these datasets showing their format as: sentence, targeted expression and the provided label; 2) the statistics of these datasets including their size and percentage of metaphors. 

\paragraph{Relation-Level Datasets:} These datasets focus on expressions of certain grammatical relations. Obtaining these relations could be done either automatically by employing a dependency parser or manually by highlighting targeted expressions in a specific corpus. Then, these expressions are manually annotated for metaphoricity given the surrounding context. There exist two benchmark datasets of this kind, namely the \textbf{TSV} dataset and \citet{Zayed:2019} \textbf{Tweets} dataset, hereafter \textbf{ZayTw} dataset. The former focuses on discerning the metaphoricity of adjective-noun expressions in sentences collected from the Web and Twitter while the latter focuses on verb-direct object expressions in tweets.

\paragraph{Adapted Word-Level Datasets:} Annotated datasets that support word-level metaphor identification are not suitable to support relation-level processing due to the annotation difference~\cite{Shutova:2015,Zayed:2020}. To overcome the limited availability of relation-level datasets, there has been a growing effort to enrich and extend benchmark datasets annotated on the word-level to suit relation-level metaphor identification. Although it is non-trivial and requires extra annotation effort, \citet{Shutova:2016} and \citet{Zayed:2020} introduced adapted versions of the MOH, TroFi and VUAMC datasets to train and evaluate models that identify metaphors on the relation-level. 
Since the \textbf{MOH} dataset was originally created to identify metaphoric verbs on the word-level, its adaptation by~\citet{Shutova:2016}, also referred to as \textbf{MOH-X} in several papers, focused on extracting the verb-noun grammar relations using a dependency parser. The dataset is relatively small and contains short and simple sentences that are originally sampled from the example sentences of each verb in WordNet~\cite{Fellbaum:1998}. The \textbf{TroFi} dataset was designed to identify the metaphoricity of 50 selected verbs on the word-level from the 1987-1989 Wall Street Journal (WSJ) corpus. The \textbf{VUAMC}~\cite{Steen:2010} is the largest corpus annotated for metaphors and has been employed extensively by models developed to identify metaphors on the word-level. However, models designed to support relation-level metaphor identification can not use it in its current state. Therefore, previous research focusing on relation-level processing \cite{rei:2017,bulat:2017,Shutova:2016,Tsvetkov:2014} did not train, evaluate or compare their approaches using it. Recently, a subset of the VUAMC was adapted to suit relation-level analysis by focusing on the training and test splits provided by the NAACL metaphor shared task. 
This corpus subset as well as the TroFi dataset are adapted by~\citet{Zayed:2020} to suit identifying metaphoric expressions on the relation-level, focusing on verb-direct object grammar relations (i.e \textit{dobj} dependencies). The Stanford dependency parser was utilised to extract these relations which were then filtered to ensure quality.

\section{Experiments}
\label{sec:experiments}

\subsection{Experimental Setup}
We employ a single-layer LSTM model with 512 hidden units. The Adadelta algorithm~\cite{zeiler:2012_adadelta} is used for optimisation during the training phase and the binary cross-entropy is used as a loss function to fine tune the network. The reported results are obtained using batch size of 256 instances for the ZayTw dataset and 128 instances for the other employed datasets. ${L_2}$-regularisation weight of 0.01 is used to constraint the weights of the contextual modulator. In all experiments, we zero-pad the input sentences to the longest sentence length in the dataset. All the hyper-parameters were optimised on a randomly separated development set (validation set) by assessing the accuracy. We present here the best performing design choices based on experimental results but we highlight some other attempted considerations in Appendix~\ref{sec:appendix_b}. We implemented our models using Keras~\cite{chollet:2015:keras} with the TensorFlow backend. We are making the source code and best models publicly available\footnote{\url{https://github.com/OmniaZayed/affineTrans_metaphor_identification}}. To ensure reproducibility, we include the sizes of the training, validation and test sets in Appendix~\ref{sec:appendix_b} as well as the best validation accuracy obtained on each validation set. All the results presented in this paper are obtained after running the experiments five times with different random seeds and taking the average. 

In this work, we selected the following state-of-the-art models pertaining to relation-level metaphor identification for comparisons: the cross-lingual model by~\cite{Tsvetkov:2014}, the multimodal system of linguistic and visual features by~\cite{Shutova:2016}, the ATTR-EMBED model by~\citet{bulat:2017} and the supervised similarity network (SSN) by~\citet{rei:2017}. We consider the SSN system as our baseline. For fair comparisons, we utilised their same data splits on the five employed benchmark datasets described in Section~\ref{sec:datasets}.

\subsection{Excluding \textit{AffineTrans}}
\label{subsec:baseline}
We implemented a simple LSTM model to study the effect of employing affine transformations on the system performance. The input to this model is the tokenised sentence $S$ which is embedded as a sequence of vector representations using GloVe. These sequences of word embeddings are then encoded using the LSTM layer to compute a contextual representation. Finally, this representation is fed to a feed-forward layer with a sigmoid activation to predict the class label. We used this model with and without the attention mechanism.

\subsection{Results}
\label{sub-sec:results}
We conduct several experiments to better understand our proposed model. First, we experiment with the simple model introduced in Section~\ref{subsec:baseline}. Then, we train the proposed models on the benchmark datasets discussed in Section~\ref{sec:datasets}. We experiment with and without the attention layer to assess its effect on the model performance. Furthermore, we compare our model to the current work that addresses the task on the relation-level, in-line with our peers in this area. Tables~\ref{tab:results_compare_1} and~\ref{tab:results_compare_2} show our model performance in terms of precision, recall, F1-score and accuracy. 

Since the source code of Rei's~\citeyearpar{rei:2017} system is available online\footnote{\url{https://github.com/marekrei/ssn}}, we trained and tested their model using the ZayTw dataset as well as the adapted VUAMC and TroFi dataset in an attempt to study the ability of their model to generalise when applied on a corpus of a different text genre with wider metaphoric coverage including less common (conventionalised) metaphors.

\begin{table*}[!htb]
\centering
	\resizebox{0.9\textwidth}{!}{
		\begin{tabular}{l|llll|llll}
			\hline \specialrule{2pt}{1pt}{0pt}
			  & \multicolumn{4}{c|}{\textbf{ZayTw (test-set)}}  &  \multicolumn{4}{c}{\textbf{TSV (test-set)}}   \\
			 &  \textbf{Prec.}  &  \textbf{Recall}  &  \textbf{F1-score}  &  \textbf{Acc.}  &  \textbf{Prec.}  &  \textbf{Recall}  &  \textbf{F1-score}  &  \textbf{Acc.}  \\ \hline
			
			\citet{Tsvetkov:2014} & - & - & - & - & - & - & \textbf{0.85} & - \\
			 \citet{Shutova:2016} (multimodal)  &  -  &  -  &  -  &  -  &  0.67 & \textbf{0.96} & 0.79 &  -   \\
			 \citet{bulat:2017} (ATTR-EMBED) & - & - & - & - & 0.85 & 0.71 & 0.77 & - \\
			 \citet{rei:2017} (SSN)  &  0.543 &  \textbf{1.0} &  0.704 &  0.543 &  \textbf{0.903} & 0.738 & 0.811 & 0.829  \\ \hline

			Simple LSTM   & 0.625 & 0.758 & 0.685 & 0.621 & 0.690 &	0.58 & 0.630	& 0.66   \\
			Simple LSTM (+ Attend)  & 0.614	& 0.866 & 0.718 & 0.631 & 0.655 &	0.55 & 0.598 &	0.63  \\

		    Our AffineTrans & \textbf{0.804} & 0.769 & \textbf{0.786*} & \textbf{0.773} & 0.869 & 0.80	& \textbf{0.834}	& \textbf{0.84}  \\
		    Our AffineTrans (+ Attend) & 0.758 & \textbf{0.812} & 0.784* & 0.757 &  0.875 & 0.77 & 0.819 & 0.83 \\
		
		\specialrule{2pt}{0pt}{1pt} \hline 
		\end{tabular}
	}
	\caption{\label{tab:results_compare_1} Our proposed architecture performance compared to the state-of-the-art approaches on the benchmark datasets ZayTw and TSV. *Statistically significant ($p$-value$<$0.01) compared to the SSN system~\cite{rei:2017}.}
\end{table*}

\begin{table*}[!htbp]
\centering
	\resizebox{\textwidth}{!}{
		\begin{tabular}{l|llll|llll|llll}
			\hline \specialrule{2pt}{1pt}{0pt}
			  & \multicolumn{4}{c|}{\textbf{adapted MOH (10-fold)}}  &  \multicolumn{4}{c|}{\textbf{adapted TroFi (test-set)}} & \multicolumn{4}{c}{\textbf{adapted VUAMC (test-set)}}   \\
			 &  \textbf{Prec.}  &  \textbf{Recall}  &  \textbf{F1-score}  &  \textbf{Acc.}  &  \textbf{Prec.}  &  \textbf{Recall}  &  \textbf{F1-score}  &  \textbf{Acc.} &  \textbf{Prec.}  &  \textbf{Recall}  &  \textbf{F1-score}  &  \textbf{Acc.} \\ \hline
			
			 \citet{rei:2017} (SSN)  &  0.736 &  0.761 &  0.742 &  0.748 & 0.620 & 0.892 & 0.732 & 0.628 &  0.475 & 0.532 & 0.502 & 0.558 \\ \hline
			
			Simple LSTM   &  0.757 & 0.773 & 0.759 & 0.759 & 0.70 &	0.751 & 0.725 & 0.674  & 0.510	& 0.339	& 0.407	& 0.587\\
			Simple LSTM (+ Attend)  & 0.746 & 0.782 & 0.757 & 0.752  & 0.759 & 0.853	& 0.803*	& 0.761 & 0.575 & 0.423 &	0.487	& 0.627  \\
			
		    Our AffineTrans  &\textbf{0.804} & 0.748 & 0.771 & \textbf{0.780}  & \textbf{0.852} &	\textbf{0.909} &	\textbf{0.879*} &	\textbf{0.858} & \textbf{0.712} & 0.639	& 0.673*	& \textbf{0.741}  \\
		    Our AffineTrans (+ Attend) & 0.753 & \textbf{0.813} & \textbf{0.779} & 0.773  &  0.841 &	0.870 &	0.856* & 0.832 & 0.686 &	\textbf{0.679} &	\textbf{0.683*} &	0.736  \\
		
		\specialrule{2pt}{0pt}{1pt} \hline 
		\end{tabular}
	}
	\caption{\label{tab:results_compare_2} Our proposed architecture performance compared to the state-of-the-art approaches on the adapted benchmark datasets MOH, TroFi and VUAMC. *Statistically significant ($p$-value$<$0.01) compared to the SSN system~\cite{rei:2017}. We could not include~\citet{Shutova:2016} results on the MOH dataset since they used different test settings, thus their results will not be strictly comparable.}
\end{table*}

\section{Discussion}
\label{sec:discussion}

\begin{description}[style=unboxed,topsep=1pt,itemsep=0pt,leftmargin=0pt]
\item[\textbf{Overall performance.}] We analysed the model performance by inspecting the classified instances. We noticed that it did a good job identifying conventionalised metaphors as well as uncommon ones. Appendix~\ref{sec:appendix_a} shows examples of classified instances by our system from the employed benchmark datasets.
Our model achieves significantly better F1-score over the state-of-the-art SSN system~\cite{rei:2017} under the one-tailed paired \textit{t-test}~\cite{yeh:2000:onetail} at $p$-value$<$0.01 on three of the five employed benchmark datasets. Moreover, our architecture showed improved performance over the state-of-the-art approaches on the TSV and MOH datasets. It is worth mentioning that the size of their test sets is relatively smaller; therefore any change in a single annotated instance drastically affects the results. Moreover, the approach proposed by \citet{Tsvetkov:2014} relies on hand-coded lexical features which justifies its high F1-score.

\item[\textbf{The effect of contextual modulation.}] When excluding the \textit{AffineTrans} layer and only using the simple LSTM model, we observe a significant performance drop that shows the effectiveness of leveraging linear modulation. This layer adaptively influences the output of the model by conditioning the identification process on the contextual information of the targeted expression itself which significantly improved the system performance, as observed from the results. Moreover, employing the contextualised representation of the targeted expression, through ELMo sentence embeddings, was essential to explicitly capture the interaction between the verb/adjective and its accompanying noun. Then, the \textit{AffineTrans} layer was able to modulate the network based on this interaction.

\item[\textbf{The effect of attention.}] It is worth noting that the attention mechanism did not help much in our \textit{AffineTrans} model because affine transformation itself could be seen as playing a similar role to attention, as discussed in Section~\ref{sec:intro}. In attention mechanisms important elements are given higher weight based on weight scaling whereas in linear affine transformation scaling is done in addition to shifting which gives prior importance (probability) to particular features. We are planning to perform an in-depth comparison of using affine transformation verses attention in our future work.

\item[\textbf{Error analysis.}] An error analysis is performed to determine the model flaws by analysing the predicted classification. We examined the false positives and false negatives obtained by the best performing model, namely \textit{AffineTrans} (without attention). Interestingly, the majority of false negatives are from the political tweets in ZayTw dataset. Table~\ref{tab:analysis} lists some examples of misclassified instances in the TSV and ZayTw datasets. Some instances could be argued as being correctly classified by the model. For instance, \textit{``spend capital''} could be seen as a metaphor in that the noun is an abstract concept referring to actual money. Examples of misclassified instances from the other employed datasets are presented in Appendix~\ref{sec:appendix_a}. Interestingly, we noticed that the model was able to spot mistakenly annotated instances.
Although the adapted VUMAC subset contains various expressions which should help the model perform better, we noticed annotation inconsistency in some of them. For example, the verb \textit{``choose''} associated with the noun \textit{``science''} is annotated once as metaphor and twice as literal in very similar contexts. This aligns well with the findings of~\citet{Zayed:2020} who questioned the annotation of around 5\% of the instances in this subset mainly due to annotation inconsistency.

\begin{table*}[!htb]
	\centering
		\resizebox{\textwidth}{!}{
		 \def\arraystretch{1.2}
			\begin{tabular}{l|p{7cm}|l||p{7cm}|l}
				\hline \specialrule{2pt}{1pt}{0pt}
				& \multicolumn{2}{c||}{\textbf{ZayTw}}  &  \multicolumn{2}{c}{\textbf{TSV}} \\
            &\textbf{Tweet} & \textbf{Prob.} &\textbf{Sentence} & \textbf{Prob.} \\ \hline
            
    & hard to \textbf{\textcolor{red}{resist the feeling}} that remain is further [...] & 0.46 & You have a \textbf{\textcolor{red}{shiny goal}} in mind that is distracting you with its awesomeness. & 0.49 \\ \cmidrule{2-5}
    \vtop{\hbox{\strut \textbf{False} }\hbox{\strut \textbf{Negative}}}  & @abpi uk: \textbf{\textcolor{red}{need \#euref final facts}}? read why if [...] & 0.08 & The first hours of a \textbf{\textcolor{red}{shaky ceasefire}} are not ``the best of times''. & 0.14 \\ \cmidrule{2-5}
    & \#ivoted with a black pen. do not \textbf{\textcolor{red}{trust pencils}}. [...] & 0.003 & The French bourgeoisie has rushed into a \textbf{\textcolor{red}{blind alley}}. & 0.00\\ 
    \hline
    
    & [...] this guy would \textbf{\textcolor{red}{spend so much political capital}} trying to erase the [...] & 0.96 & I could hear the \textbf{\textcolor{red}{shrill voices}} of his sisters as they dash about their store helping customers. & 0.98 \\ \cmidrule{2-5} 
    \vtop{\hbox{\strut \textbf{False} }\hbox{\strut \textbf{Positive}}} & \#pencilgate to \textbf{\textcolor{red}{justify vitriolic backlash}} if \#remain wins [...] & 0.94 & [...] flavoring used in cheese, meat and fish to give it a \textbf{\textcolor{red}{smoky flavor}} could in fact be toxic. & 0.82 \\ \cmidrule{2-5}
    & @anubhuti921 @prasannas it \textbf{\textcolor{red}{adds technology}} to worst of old police state practices, [...] & 0.76* & Usually an overly dry nose is a precursor to a \textbf{\textcolor{red}{bloody nose}}. & 0.64 \\
    
    \specialrule{2pt}{0pt}{1pt}	\hline
	\end{tabular}
	}
	\caption{\label{tab:analysis} Misclassified examples by our \textit{AffineTrans} model (without attention) from ZayTw and TSV test sets. Sentences are truncated due to space limitations. *Our model was able to spot some mistakenly annotated instances.}
\end{table*}

\item[\textbf{Analysis of some misclassified verbs.}] We noticed that sometimes the model got confused while identifying the metaphoricity of expressions where the verb is related to emotion and cognition such as: \textit{``accept, believe, discuss, explain, experience, need, recognise,} and \textit{want''}. Our model tends to classify them as not metaphors. We include different examples from the ZayTw dataset of the verbs \textit{``experience''} and \textit{``explain''} with different associated nouns along with their gold and predicted classifications in Appendix~\ref{sec:appendix_a}.  Our model's prediction seems reasonable given that the instances in the training set were labelled as not metaphors. It is not clear why the gold label for \textit{``explain this mess''} is not a metaphor while it is metaphor for \textit{``explain implications''}; similarly, the nouns \textit{``insprirations''} and \textit{``emotions''} with the verb \textit{``experience''}.

\end{description}

\section{Conclusions}
\label{sec:conclusion}
In this paper, we introduced a novel architecture to identify metaphors by utilising feature-wise affine transformation and deep contextual modulation. Our approach employs a contextual modulation pipeline to capture the interaction between the metaphor components. This interaction is then used as an auxiliary input to modulate a metaphor identification linguistic pipeline. We showed that such modulation allowed the model to dynamically highlight the key contextual features to identify the metaphoricity of a given expression. We applied our approach to relation-level metaphor identification to classify expressions of certain syntactic constructions for metaphoricity as they occur in context. We significantly outperform the state-of-the-art approaches for this level of analysis on benchmark datasets. Our experiments also show that our contextual modulation-based model can generalise well to identify the metaphoricity of unseen instances in different text types including the noisy user-generated text of tweets. Our model was able to identify both conventionalised common metaphoric expressions as well as less common ones. To the best of our knowledge, this is the first attempt to computationally identify metaphors in tweets and the first approach to study the employment of feature-wise linear modulation on metaphor identification in general. The proposed methodology is generic and can be applied to a wide variety of text classification approaches including sentiment analysis or term extraction.

\section*{Acknowledgments}

This work was supported by Science Foundation Ireland under grant number SFI/12/RC/2289\_2 (Insight). 

{\noindent We would like to thank the anonymous reviewers of this paper for their helpful comments and feedback. Special thanks for the anonymous meta-reviewer for steering an effective and constructive discussion about this paper which we realised its results through the experienced, extensive and beneficial meta-review. Sincere thanks to Mennatullah Siam for the insightful discussions about the technical part of this paper.}

\bibliographystyle{acl_natbib}
\bibliography{metaphor_ident_emnlp2020}

\appendix

\section{Datasets Statistics and Analysis}
\label{sec:appendix_a}

\subsection{Benchmark Datasets Statistics}
Table~\ref{tab:datasets_stat} shows the statistics of the benchmark datasets employed in this work, namely the relation-level datasets TSV\footnote{\url{https://github.com/ytsvetko/metaphor}} and ZayTw in addition to the adapted TroFi\footnote{\url{http://natlang.cs.sfu.ca/software/trofi.html}}, VUAMC\footnote{\url{http://ota.ahds.ac.uk/headers/2541.xml}} and MOH\footnote{\url{http://saifmohammad.com/WebPages/metaphor.html}} datasets. Table~\ref{tab:datasets_examples} shows examples of annotated instances from each dataset.

\begin{table*}[!htb]
	\resizebox{\textwidth}{!}{
		\def\arraystretch{1.3}
		
		\begin{tabular}{l|l|l|l|l|l}
			\hline \specialrule{2pt}{1pt}{0pt}
			\textbf{Dataset}  & \textbf{Syntactic structure} & \textbf{Text type} & \textbf{Size} & \textbf{\% Metaphors} & \vtop{\hbox{\strut \textbf{ Average} }\hbox{\strut \textbf{ Sentence}}\hbox{\strut \textbf{Length}}}\\ \hline
			 
			The adapted TroFi Dataset & verb-direct object & \vtop{\hbox{\strut 50 selected verbs}\hbox{\strut (News)}}& 1,535 sentences  & 59.15\% & 48.5 \\ 
			 
			 \vtop{\hbox{\strut The adapted VUAMC }\hbox{\strut (NAACL Shared Task subset)}} & verb-direct object & \vtop{\hbox{\strut known-corpus}\hbox{\strut (The BNC)}} &  5,820 sentences & 38.87\% & 63.5 \\
			 
			The adapted MOH Dataset & \vtop{\hbox{\strut verb-direct object;}\hbox{\strut subject-verb}} & \vtop{\hbox{\strut selected examples}\hbox{\strut (WordNet)}} & 647 sentences & 48.8\% & 11 \\
            
            The TSV Dataset & adjective–noun & \vtop{\hbox{\strut selected examples}\hbox{\strut (Web/Tweets)}} & 1,964 sentences & 50\% & 43.5  \\ 
			 
            The ZayTw Dataset & verb-direct object & \vtop{\hbox{\strut Tweets}\hbox{\strut (general and political topics)}} & 2,531 tweets & 54.8\% & 34.5 \\ 
		\specialrule{2pt}{0pt}{1pt} \hline 
		\end{tabular}
	}
	\caption{\label{tab:datasets_stat} Statistics of the employed benchmark datasets to train and evaluate our proposed models highlighting the used experimental setting and links to the data sources in the footnotes. The adapted versions are available upon request from their corresponding authors.}
\end{table*}

\begin{table*}[!htbp]
	\centering
		\resizebox{\textwidth}{!}{
		 \newcommand*\rot{ \centering \rotatebox{90}}

			\begin{tabular}{l|p{9cm}|l|l}
				\hline \specialrule{2pt}{1pt}{0pt}
            \textbf{Dataset} & \textbf{Sentence} & \textbf{Targeted Expression} & \vtop{\hbox{\strut \textbf{Gold}}\hbox{\strut \textbf{Label}}} \\ \hline
         
     & Chicago is a big city, with a lot of everything to offer. & 	big city	& 0 \\ \cmidrule{2-4}
     \multirow{4}{*}{\rot{\textbf{TSV}}} & It 's a foggy night and there are a lot of cars on the motorway. 	& foggy night	& 0\\ \cmidrule{2-4}
     & Their initial icy glares had turned to restless agitation. & icy glares & 1\\ \cmidrule{2-4}
     & And he died with a sweet smile on his lip. & sweet smile & 1\\ \hline
     
      & insanity. ok to abuse children by locking them in closet, dark room and damage their psyche, but corporal punishment not ok? twisted! & abuse children & 0 \\  \cmidrule{2-4}
     \multirow{4}{*}{\rot{\textbf{ZayTw}}} & nothing to do with your lot mate \#ukip ran hate nothing else and your bloody poster upset the majority of the country regardless in or out	& upset the majority &	0 \\ \cmidrule{2-4}
     & nothing breaks my heart more than seeing a person looking into the mirror with anger \& disappointment, blaming themselves when someone left.	& breaks my heart & 1 \\\cmidrule{2-4}
     & how quickly will the warring tories patch up their differences to preserve power? \#euref & patch up their differences & 1\\  \hline
    
    &  A Middle Eastern analyst says Lebanese usually drink coffee at such occasions; Palestinians drink tea. &	drink coffee & 0\\ \cmidrule{2-4}
    \multirow{4}{*}{\rot{ \textbf{ The adapted TroFi}}}   & In addition, the eight-warhead missiles carry guidance systems allowing them to strike Soviet targets precisely. &	strike Soviet targets	 & 0\\ \cmidrule{2-4}
     & He now says that specialty retailing fills the bill, but he made a number of profitable forays in the meantime. &	fills the bill	& 1\\ \cmidrule{2-4}
    & A survey of U.K. institutional fund managers found most expect London stocks to be flat after the fiscal 1989 budget is announced, as Chancellor of the Exchequer Nigel Lawson strikes a careful balance between cutting taxes and not overstimulating the economy. &	strikes a careful balance	& 1  \\ \hline
    
     & Among the rich and famous who had come to the salon to have their hair cut, tinted and set, Paula recognised Dusty Springfield, the pop singer, her eyes big and sooty , her lips pearly pink, and was unable to suppress the thrill of excitement which ran through her. & recognised Dusty Springfield	& 0   \\ \cmidrule{2-4}
     \multirow{4}{*}{\rot{\vtop{\hbox{\strut \textbf{The adapted VUAMC}}\hbox{\strut (NAACL Shared Task)}}}}  & But until they get any money back, the Tysons find themselves in the position of the gambler who gambled all and lost . & get any money & 0\\ \cmidrule{2-4}
     & The Labour Party Conference: Policy review throws a spanner in the Whitehall machinery & throws a spanner & 1 \\ \cmidrule{2-4}
    & Otherwise Congress would have to face the consequences of automatic across-the-board cuts under the Gramm-Rudman-Hollings budget deficit reduction law. & face the consequences & 1 \\ \hline
    
     \multirow{4}{*}{\rot{\textbf{MOH-X}}}  &  commit a random act of kindness. & commit a random act & 0 \\ \cmidrule{2-4}
     & The smoke clouded above the houses.	& smoke clouded & 0 \\ \cmidrule{2-4}
     & His political ideas color his lectures.	& ideas color & 1\\ \cmidrule{2-4}
     & flood the market with tennis shoes. & flood the market & 1 \\ 
   
     \specialrule{2pt}{0pt}{1pt}	\hline
	\end{tabular}
	}
	\caption{\label{tab:datasets_examples} Examples of annotated instances from the employed relation-level datasets showing their format as: sentence, targeted expression and the provided label. }
\end{table*}

\subsection{Datasets Analysis}

\paragraph{Examples of correctly classified instances from the employed datasets:} We show examples of correctly classified instances by our best performing model. Table~\ref{tab:classified_instances_1} comprises examples from the relational-level datasets TSV and ZayTw. Table~\ref{tab:classified_instances_2} lists examples from the adapted MOH and TroFi datasets as well as the adapted VUAMC.


\begin{table*}[!htb]
\centering
	\resizebox{\textwidth}{!}{
		\def\arraystretch{1.2}
		\begin{tabular}{l|p{6cm}l||p{5cm}l}
			\hline \specialrule{2pt}{2pt}{0pt}
			  \textbf{Model} &\multicolumn{2}{c||}{ \textbf{ZayTw}} & \multicolumn{2}{c}{ \textbf{TSV}}\\ 
			\textbf{Classification} &\textbf{Expression} & \textbf{Prob.} & \textbf{Expression} & \textbf{Prob.} \\
			\hline
			& poisoning our democracy & 0.999 &  rich history & 0.999  \\ 
			& binding the country & 0.942 &  rocky beginning & 0.928 \\
			\textbf{Metaphor} & see greater diversity & 0.892 & foggy brain & 0.873 \\
			& patch up their differences & 0.738 & steep discounts & 0.723 \\
			& seeking information & 0.629 & smooth operation & 0.624 \\
			& retain eu protection & 0.515 & dumb luck & 0.512 \\ \hline
		    & shake your baby& 0.420 & filthy garments & 0.393 \\
			& enjoy a better climate& 0.375 & clear day & 0.283\\
			\textbf{Not Metaphor} & improve our cultural relations& 0.292 & slimy slugs & 0.188 \\
			& placate exiters & 0.225 & sour cherries & 0.102 \\
			& betrayed the people & 0.001 & short walk & 0.014 \\
			& washing my car & 0.000 & hot chocolate & 0.000 \\
			\specialrule{2pt}{0pt}{1pt}	\hline
		\end{tabular}
	}
\caption{\label{tab:classified_instances_1} Examples of correctly classified instances by our \textit{AffineTrans} model (without attention) from the ZayTw and TSV datasets showing the classification probability.}
\end{table*}

\begin{table*}[!htbp]
\centering
	\resizebox{\textwidth}{!}{
		\def\arraystretch{1.2}
		\begin{tabular}{l|ll||ll||ll}
			\hline \specialrule{2pt}{2pt}{0pt}
			  \textbf{Model} &\multicolumn{2}{c||}{ \textbf{adapted MOH}} & \multicolumn{2}{c||}{ \textbf{adapted TroFi}} & \multicolumn{2}{c}{ \textbf{adapted VUAMC}}\\ 
			\textbf{Classification} &\textbf{Expression} & \textbf{Prob.} & \textbf{Expression} & \textbf{Prob.}  & \textbf{Expression} & \textbf{Prob.} \\
			\hline
			& absorbed the knowledge & 0.987 & grasped the concept & 0.985 & bury their reservations & 0.999 \\ 
			& steamed the young man & 0.899  & strike fear & 0.852 & reinforce emotional reticence & 0.871\\
			\textbf{Metaphor} & twist my words & 0.770  & ate the rule & 0.781 & possess few doubts & 0.797 \\
		    & color my judgment & 0.701  & planted a sign & 0.700 & suppress the thrill	& 0.647 \\
			& poses an interesting question & 0.543  & examined the legacy & 0.599 & considers the overall effect & 0.568 \\
			& wears a smile & 0.522 & pumping money & 0.529 & made no attempt & 0.517 \\ \hline
		    & shed a lot of tears & 0.484  & pumping power & 0.427 & send the tape & 0.482 \\
			& abused the policeman & 0.361  & poured acid & 0.314 & asking pupils & 0.389 \\
			\textbf{Not Metaphor} &  tack the notice & 0.274  & ride his donkey & 0.268 & removes her hat & 0.276 \\
			& stagnate the waters & 0.148  & fixed the dish & 0.144 & enjoying the reflected glory & 0.188 \\
			& paste the sign & 0.002  & lending the credit & 0.069 & predict the future	& 0.088\\
			& heap the platter & 0.000  & destroy coral reefs & 0.000& want anything & 0.000\\
			\specialrule{2pt}{0pt}{1pt}	\hline
		\end{tabular}
	}
\caption{\label{tab:classified_instances_2} Examples of correctly classified instances by our \textit{AffineTrans} model (without attention) from the adapted MOH, TroFi and VUAMC datasets showing the classification probability.}
\end{table*}

\paragraph{Examples of misclassified instances by our model in the tweets dataset:} Examples of misclassified instances from the TSV and ZayTw datasets as well as the adapted MOH, TroFi and VUAMC datasets are given in Table~\ref{tab:misclassified_2}. Our model spotted some instances that are mistakenly annotated in the original datasets.

\begin{table*}[!htbp]
	\centering
		\resizebox{\textwidth}{!}{
		\newcommand*\rot{ \centering \rotatebox{90}}
			\begin{tabular}{l|l|p{12cm}|l}
				\hline \specialrule{2pt}{1pt}{0pt}
            &\textbf{Dataset} & \textbf{Sentence} & \textbf{Prob.} \\ \hline

    & \textbf{TroFi} & Unself-consciously , the littlest cast member with the big voice steps into the audience in one number to open her wide cat-eyes and throat to \textbf{\textcolor{red}{melt the heart}} of one lucky patron each night. & 0.295 \\ \cmidrule{3-4}
    & & Lillian Vernon Corp., a mail-order company, said it is experiencing delays in \textbf{\textcolor{red}{filling orders}} at its new national distribution center in Virginia Beach,Va.& 0.006\\ \cmidrule{2-4}
    
    \multirow{10}{*}{\rot{\textbf{False Negative}}}  & \textbf{VUAMC} &  It is a curiously paradoxical foundation uponupon which to \textbf{\textcolor{red}{build a theory}} of autonomy. & 0.410 \\ \cmidrule{3-4} 
    & &  It has turned up in Canberra with Japan to develop Asia Pacific Economic Co-operation (APEC) and a new 12-nation organisation which will \textbf{\textcolor{red}{mimic the role}} of the Organisation for Economic Co-operation and Development in Europe. & 0.000 \\ \cmidrule{2-4}
    
     & \textbf{MOH} & When does the \textbf{\textcolor{red}{court of law sit}}? & 0.499 \\ \cmidrule{3-4}
    & & The \textbf{\textcolor{red}{rooms communicated}}. & 0.000 \\ \cmidrule{2-4}
    
    & \textbf{TSV} & It was great to see a \textbf{\textcolor{red}{warm reception}} for it on twitter. & 0.488  \\ \cmidrule{3-4}
    & &  An \textbf{\textcolor{red}{honest meal}} at a reasonable price is a rarity in Milan. & 0.000 \\ \cmidrule{2-4}
     
     & \textbf{ZayTw} & \#brexit? we \textbf{\textcolor{red}{explain likely implications}} for business insurances on topic of \#eureferendum & 0.2863 \\ \cmidrule{3-4}
     & & @abpi uk: \textbf{\textcolor{red}{need \#euref final facts}}? read why if you care about uk life sciences we're \#strongerin. & 0.0797 \\ \hline

    & \textbf{TroFi} & As the struggle enters its final weekend , any one of the top contenders could \textbf{\textcolor{red}{grasp his way}} to the top of the greasy pole. & 0.998* \\ \cmidrule{3-4}
    & & Southeastern poultry producers fear \textbf{\textcolor{red}{withering soybean supplies}} will force up prices on other commodities. & 0.507 \\ \cmidrule{2-4}
    
    \multirow{10}{*}{\rot{\textbf{False Positive}}} & \textbf{VUAMC} & Or after we \textbf{\textcolor{red}{followed the duff advice}} of a legal journalist in a newspaper? & 0.999* \\ \cmidrule{3-4}
    & &  Aristotle said something very interesting in that extract from the Politics which I quoted earlier; he said that women have a deliberative faculty but that it \textbf{\textcolor{red}{lacks full authority}}. & 0.525 \\ \cmidrule{2-4}  
    
    & \textbf{MOH} &  All our \textbf{\textcolor{red}{planets condensed}} out of the same material. & 0.999 \\ \cmidrule{3-4} 
    & & \textbf{\textcolor{red}{He bowed}} before the King. & 0.868 \\ \cmidrule{2-4}
    
    & \textbf{TSV} &  Bags two and three will only have \textbf{\textcolor{red}{straight edges}} along the top and the bottom. &  0.846 \\  \cmidrule{3-4}
    & &  Mountain climbers at \textbf{\textcolor{red}{high altitudes}} quickly acquire a tan from the sun. & 0.986\\ \cmidrule{2-4}
    
    & \textbf{ZayTw} & delayed flight in fueturventura due to french strikes \textbf{\textcolor{red}{restricting access}} across french airspace =/ hopefully get back in time to \#voteleave & 0.9589\\ \cmidrule{3-4}
    & & in manchester more young people are expected to \textbf{\textcolor{red}{seek help}} in the coming months and years \#cypiapt \#mentalhealth & 0.7055* \\  
    
     \specialrule{2pt}{0pt}{1pt}	\hline
	\end{tabular}
	}
	\caption{\label{tab:misclassified_2} Misclassified examples by our \textit{AffineTrans} model (without attention) from the TSV test set as well as the adapted MOH, TroFi and VUAMC test sets. *Our model was able to spot some mistakenly annotated instances in the dataset.}
\end{table*}

\paragraph{Missclassified Verbs:} Table~\ref{tab:verbs_examples} shows examples from the ZayTw dataset of the verbs \textit{``experience''} and \textit{``explain''} with different associated nouns along with their gold and predicted classifications.

\begin{table*}[!htbp]
	\resizebox{\textwidth}{!}{
	\def\arraystretch{1.2}
	\newcommand*\rot{ \centering \rotatebox{90}}
		\begin{tabular}{llp{8.5cm}lll}
			\hline \specialrule{2pt}{1pt}{0pt}
			 \multicolumn{2}{c}{\textbf{Expression}}  & tweet & \textbf{Predicted}  &  \textbf{Prob.}  &  \textbf{Gold}  \\ \hline
			 & the inspiration & relive the show , re - listen to her messages, re - experience the inspiration, refuel your motivation & \textbf{\textcolor{red}{0}} & 0.220 & 1 \\
            \multirow{3}{*}{\rot{\textbf{experience}}} & your emotions & do not be afraid to experience your emotions; they are the path to your soul. trust yourself enough to feel what you feel. & 0 & 0.355 & 0 \\
            & this shocking behaviour & a friend voted this morning \& experienced this shocking behaviour. voting is everyone 's right. \#voteremain & 0 & 0.009 & 0 \\ \hline
            
            & likely implications & \#brexit? we explain likely implications for business insurances on topic of \#eureferendum &\textbf{\textcolor{red}{0}} & 0.2866 & 1 \\
            \multirow{3}{*}{\rot{\textbf{explain}}} &  this mess & @b\_hanbin28 ikr same here :D imagine hansol \& shua trynna explain this mess to other members :D & 0 & 0.109 & 0 \\
            & the rise & loss aversion partly explains the rise of trump and ukip & 1 & 0.618 & 1 \\
		\specialrule{2pt}{0pt}{1pt} \hline 
		\end{tabular}
	}
	\caption{\label{tab:verbs_examples} Examples of classified instances of the verbs \textit{``experience''} and \textit{``explain''} in the ZayTw test set. }
\end{table*}

\section{Design Considerations}
\label{sec:appendix_b}

\subsection{Experimental Settings}
The word embeddings layer is intialised with the pre-trained GloVe embeddings.  We used the uncased 200-dimensional GloVe embeddings pre-trained on $\sim$2 billion tweets and contains 1.2 million words. We did not update the weights of these embeddings during training. Table~\ref{tab:datasets_setup} shows the sizes of the training, validation and test sets of each employed dataset for as well as the corresponding best obtained validation accuracy by the the \textit{AffineTrans} model (without attention). All experiments are done on a NVIDIA Quadro M2000M GPU and the average running time for the proposed models is around 1 hour for maximum of 100 epochs.

\begin{table*}[!htbp]
	\resizebox{\textwidth}{!}{
		\def\arraystretch{1.3}
		\begin{tabular}{l|l|l|l|l|l|l}
			\hline \specialrule{2pt}{1pt}{0pt}
			\textbf{Dataset}  & \textbf{Train} & \textbf{Validation} & \textbf{Test} & \textbf{split \%} & \vtop{\hbox{\strut \textbf{ Validation}}\hbox{\strut \textbf{Accuracy}}} & \textbf{@epoch} \\ \hline
			 
			The adapted TroFi Dataset & 1,074 & 150 & 312 & 70-10-20 & 0.914 & 40 \\ 
			 
			The adapted VUAMC & 3,535 & 885 & 1,398 & - & 0.748 & 20
  \\
			 
			The adapted MOH Dataset & 582 per fold & - & 65 per fold & 10-fold cross-validation & - & - \\
            
            The TSV Dataset & 1,566 & 200 & 200 & - & 0.905 & 68 \\ 
			 
            The ZayTw Dataset & 1,661 & 360 & 510 & 70-10-20 & 0.808 & 29 \\ 
		\specialrule{2pt}{0pt}{1pt} \hline 
		\end{tabular}
	}
	\caption{\label{tab:datasets_setup} Experimental information of the five benchmark datasets including the best obtained validation accuracy by the \textit{AffineTrans} model (without attention). We preserved the splits used in literature for the VUAMC and TSV datasets.}
\end{table*}

\subsection{Other Trials}
\paragraph{Sentence Embedding:} We experimented with different representations other than GLoVe in order to embed the input sentence. We tried to employ the contextualised pre-trained embeddings ELMo and BERT  either instead of the GloVe embeddings or as additional-features but no further improvements were observed on both validation and test sets over the best performance obtained. Furthermore, we experimented with different pre-trained GloVe embeddings including the uncased 300-dimensional pre-trained vectors on the Common Crawl dataset but we did notice any significant improvements.

\paragraph{Sentence Encoding:} The choice of using the simple LSTM to encode the input was based on several experiments on the validation set. We tried bidirectional LSTM but observed no further improvement. This is due to the nature of the relation-level metaphor identification task itself as the tenor (e.g. noun) affects the metaphoricity of the vehicle (e.g. verb or adjective) so a single-direction processing was enough.

\end{document}